\documentclass{article}
\usepackage{iclr2027_conference,times}

\usepackage{amsmath,amsfonts,bm}

\def\eqref#1{equation~\ref{#1}}

\def\1{\bm{1}}

\DeclareMathAlphabet{\mathsfit}{\encodingdefault}{\sfdefault}{m}{sl}
\SetMathAlphabet{\mathsfit}{bold}{\encodingdefault}{\sfdefault}{bx}{n}

\usepackage{amsmath,amssymb,mathtools,graphicx,booktabs,array,multirow,xcolor,xspace}
\usepackage{microtype,enumitem,tikz,placeins,adjustbox,float,needspace}
\usetikzlibrary{positioning,fit,arrows.meta,calc}
\usepackage{hyperref,url,xurl}
\hypersetup{pdftitle={SetOPD: From Few Visual Exemplars to Multimodal Candidate Sets for Remote-Sensing Open-Prompt Detection},pdfauthor={Jinlong Hu and Yi Zhang and Zhiqi Xia and Yikang Zhou and Shunping Ji}}
\graphicspath{{figures/}}

\title{SetOPD: From Few Visual Exemplars to Multimodal Candidate Sets\\for Remote-Sensing Open-Prompt Detection}

\author{Jinlong Hu$^{1}$ \And Yi Zhang$^{2}$ \And Zhiqi Xia$^{1}$ \And Yikang Zhou$^{1}$ \And Shunping Ji$^{1}$\\
$^{1}$Wuhan University\\
$^{2}$Institute of Seismology, China Earthquake Administration}

\newcommand{\pqa}{PQA\xspace}
\newcommand{\br}{Base--Residual\xspace}
\newcommand{\setopd}{SetOPD\xspace}
\newcommand{\sg}{\operatorname{sg}}
\newcommand{\LN}{\operatorname{LN}}
\newcommand{\MLP}{\operatorname{MLP}}
\newcommand{\FFN}{\operatorname{FFN}}
\newcommand{\Head}{\operatorname{Head}}
\newcommand{\logitfn}{\operatorname{logit}}
\newcommand{\GIoU}{\operatorname{GIoU}}
\newcommand{\MHA}{\operatorname{MHA}}

\begin{document}
\iclrfinalcopy
\maketitle
\lhead{Preprint. Under review at ICLR 2027}

\begin{abstract}
Open-prompt detection (OPD) allows users to specify targets with text, visual exemplars, or both. Existing multimodal OPD typically compresses multiple exemplars into a class-level prototype and combines text and vision in prompt or representation space, leaving modality-specific candidate states unavailable for explicit cross-modal comparison. We formulate these two limitations from a set perspective. \setopd first introduces Base--Residual prompting, which reads boxed exemplars in scene context and maps pooled support evidence into a fixed-capacity anchor--correction visual memory that drives a dedicated visual pathway. We further introduce Paired-Query Arbitration (\pqa), in which text and visual readers decode from a shared prompt-conditioned initialization, producing naturally paired candidate states that are arbitrated within each pair and subsequently refined as a set. Across 11 heterogeneous remote-sensing sources, including four fully held-out datasets, \pqa outperforms representation-level joint prompting on every source and raises Macro11 AP from 54.00 to 55.57. It also improves over the text and Base--Residual visual readers by 1.66 and 2.02 AP, respectively, while adding only 0.63M trainable parameters. Detection-level analysis further shows that the two readers recover complementary objects and that \pqa preserves a substantial fraction of this modality-exclusive evidence.
\end{abstract}

\section{Introduction}

Open-prompt detection (OPD) allows a user to tell a detector what to find through category names, visual exemplars, or both~\citep{huang2025openrsd}. It complements open-vocabulary detection (OVD)~\citep{zareian2021ovd}: OVD asks what supervision a category received during training, whereas OPD asks through which interface a user specifies the target at inference time. This interface matters in remote sensing, where detectors are increasingly trained on unions of heterogeneous datasets with inconsistent taxonomies~\citep{pan2025laedino,huang2025openrsd}, so a base/novel split inside one dataset no longer indicates whether a concept was truly unseen and direct target specification becomes a central requirement~\citep{huang2025openrsd,yang2026rsmpod}.

\begin{figure}[t]
\centering
\includegraphics[width=\textwidth]{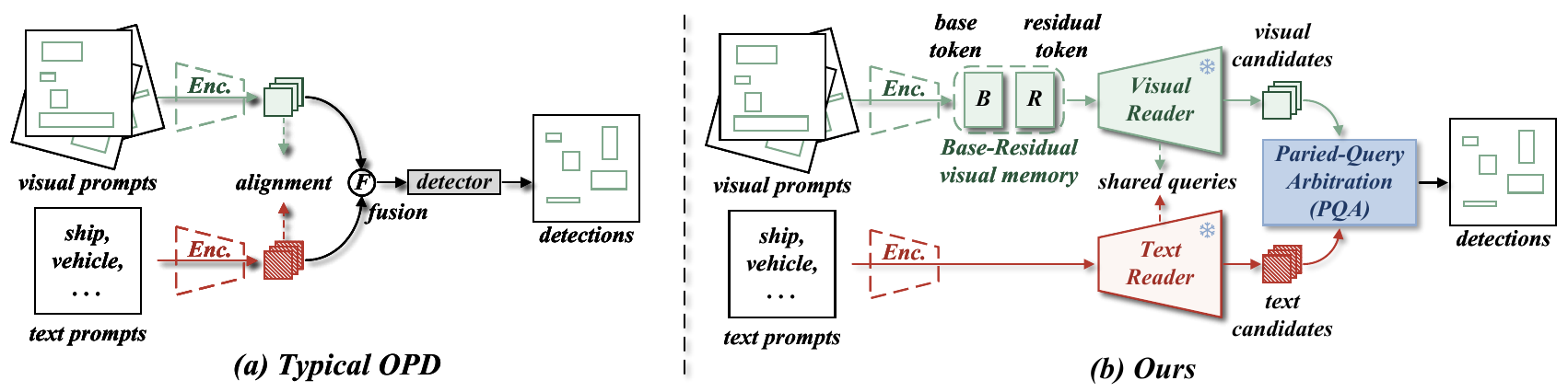}
\caption{\textbf{Typical open-prompt detectors versus \setopd.} (a)~Existing detectors compress the support set into a single class-level embedding---the visual prompt is textualized---and interact in prompt or representation space. (b)~\setopd reads a few exemplars in scene context into a fixed-capacity visual memory, lets text and visual readers decode modality-specific candidate states from a shared prompt-conditioned initialization, and combines the two modalities through candidate-state collaboration.}
\label{fig:teaser}
\end{figure}

Visual exemplars are the most direct way to specify a target: a few real instances show the detector what to look for, including appearance that a category name cannot express---which matters in remote sensing, where the same category varies substantially with scale, orientation, imaging conditions, and background. We argue that current open-prompt detectors underuse this modality in two ways, and that both are most naturally understood from a set perspective (Figure~\ref{fig:teaser}). The input to visual prompting is a support set, an unordered set of \(K\) boxed exemplars; when text is also provided, each modality can form its own set of candidate detections; and the output is itself a set, predicted by DETR-style detectors with a permutation-invariant loss~\citep{carion2020detr,zhang2023dino}.

The first issue is how a support set becomes a prompt. Exemplar-based category interfaces are well established~\citep{zang2022ovdetr,li2024dinov,jiang2024trex2,wang2025yoloe,fu2026petdino}, but given multiple exemplars, existing approaches typically compress them into a single class-level embedding through averaging, selection, or prototype aggregation~\citep{yang2026rsmpod,jiang2024trex2,qian2026detrvip}. The visual prompt is thereby textualized: one vector plays the role of a category name, often aligned with or injected into the text pathway~\citep{jiang2024trex2,xu2023mqdet,wu2025vistex}. From a set perspective, the issue is not that pooling produces a single vector---any fixed-size prompt must pool---but that the pooled vector is consumed as a category-level prototype, whose effectiveness can be sensitive to support-set size and composition in the few-example regime.

We instead keep evidence gathering visual and refuse the prototype use of its result. \br prompting reads every boxed support instance in its full scene context, so the evidence retains target position, scale, and surrounding context, and encodes any support set into a permutation-invariant prompt of \(R+1\) tokens: the pooled evidence is decomposed into a Base anchor plus a learnable Residual correction rather than used directly as a category vector, and it drives its own detection pathway rather than acting as an accessory to the text pathway. With no more than ten exemplars, this decomposition suffices: one correction raises Visual-only Macro11 AP by 6.4 points over the anchor alone, whereas routing examples into multiple competing slots brings no further gain.

The second issue is how to combine text and visual modalities. Text provides stable category semantics, while exemplars capture appearance under the current imaging conditions, but existing methods combine them early through visually enhanced language queries, aligned prompt spaces, joint multimodal prompts, or encoder/decoder-level interaction~\citep{yang2026rsmpod,jiang2024trex2,xu2023mqdet,wu2025vistex,guan2025promptdino,qin2026vitprompt}, leaving modality-specific decoded candidate states unavailable for explicit comparison and arbitration. Recent remote-sensing results also show that representation-level multimodal fusion can degrade visual prompting when textual cues are mismatched~\citep{yang2026rsmpod}. Complementarity between modalities therefore does not imply that their representations should be merged directly.

Starting from a shared prompt-conditioned proposal initialization, we preserve separate text- and visual-conditioned decoding streams and perform explicit cross-modal state interaction only after modality-specific candidate states have been formed. This turns fusion into a set problem: deciding which text candidate should be compared with which visual candidate is itself an assignment problem. Paired-Query Arbitration (\pqa) resolves it structurally---text and visual readers decode from the same initial queries and reference boxes, so the \(i\)-th candidates form a pair without any matching step, preserved under arbitrary query permutations (Proposition~2, Appendix~\ref{app:proofs}). A learned gate reads both hidden states, category confidence, and geometric disagreement to arbitrate within each pair, followed by a permutation-equivariant module that reasons over the fused candidate set. Our detection-level analysis further shows that the two readers recover non-overlapping ground-truth objects and that \pqa preserves a substantial fraction of these modality-exclusive detections across diverse sources.

In short, \setopd treats multimodal open-prompt detection as candidate-state collaboration: from a shared prompt-conditioned proposal initialization, text and visual memories produce modality-specific detection states, which \pqa arbitrates pairwise and refines jointly as a candidate set. The two components follow the same principle---retain modality-specific pathways until candidate detections are formed: exemplars are treated as visual evidence rather than category names, and text and vision are combined as two candidate sets rather than collapsed into one fused representation.

Our contributions are threefold:
\begin{itemize}[leftmargin=*,itemsep=2pt,topsep=2pt]
    \item \textbf{Set-based formulation.} We identify two structural bottlenecks in multimodal OPD---prototype-style visual prompting and representation-level cross-modal fusion---and formulate support encoding and multimodal collaboration as two set problems.
    \item \textbf{\br prompting.} We map few boxed exemplars into a fixed-capacity anchor--correction visual memory that drives a dedicated visual pathway; the default two-token representation raises Visual-only Macro11 AP from 47.11 to 53.55 over the Base-only control.
    \item \textbf{Paired-Query Arbitration.} We exploit shared initialization to pair modality-specific candidate states, arbitrate each pair, and refine the fused set; \pqa outperforms representation-level joint prompting on all 11 sources (\(+1.57\) Macro11 AP) with only 0.63M trainable parameters.
\end{itemize}

\section{Related Work}

\subsection{Open-Vocabulary and Open-Prompt Detection}

Open-vocabulary detection uses vision--language pretraining to recognize categories that do not receive bounding-box supervision~\citep{zareian2021ovd}. Language-conditioned detectors such as GLIP and Grounding DINO build transferable category interfaces through region--text alignment~\citep{li2022glip,liu2024groundingdino}, and remote-sensing detectors increasingly train on unions of heterogeneous datasets~\citep{pan2025laedino,huang2025openrsd}. Open-prompt detection instead focuses on the interface through which a target is specified. OV-DETR and OWL-ViT condition detection on image exemplars~\citep{zang2022ovdetr,minderer2022owlvit}; DINOv, T-Rex2, and YOLOE build visual or unified prompt interfaces~\citep{li2024dinov,jiang2024trex2,wang2025yoloe}; OpenRSD defines remote-sensing open-prompt detection with text and image prompts~\citep{huang2025openrsd}; and RS-MPOD combines text, visual instances, and their joint representation to specify remote-sensing targets~\citep{yang2026rsmpod}.

\subsection{Visual Prompting and Set Encoding}

Visual prompting has been extended to remote sensing, real-time unified detection, Grounding-DINO-style frameworks, and few-image open-set perception~\citep{huang2025openrsd,wang2025yoloe,fu2026petdino,zhang2025fewglances}. Given multiple exemplars, existing approaches commonly compress them by aggregation, prototype construction, or selection~\citep{yang2026rsmpod,jiang2024trex2,qian2026detrvip}. DETR-ViP further emphasizes prompt consistency, discriminability, and selective fusion~\citep{qian2026detrvip}. These methods treat exemplars as a group of objects to summarize; we instead treat them as a set whose encoding should satisfy explicit structural properties (permutation invariance and fixed capacity).

\paragraph{Set learning.}
Permutation-invariant set functions admit sum decompositions~\citep{zaheer2017deepsets}, whose simplest instance underlies prototype-based classification~\citep{snell2017proto}; learned variants include seeded attention~\citep{lee2019settransformer}, slot competition~\citep{locatello2020slot}, and residual aggregation around learnable centers~\citep{arandjelovic2016netvlad}. \br prompting reuses these building blocks but differs in three ways: it pools deviations from a support-dependent anchor rather than raw inputs or residuals from fixed centers; it targets few-shot support sets with at most ten exemplars and one to three deterministic slot queries, so the default two-token prompt is an anchor and one correction rather than competing object slots; and it is trained only through detection losses and consumed directly as a detection prompt.

\subsection{Fusing Text and Visual Evidence}

Existing multimodal detectors typically combine text and visual evidence through prompt-space alignment, visually enhanced language queries, or feature-level interaction~\citep{xu2023mqdet,jiang2024trex2,wu2025vistex,yang2026rsmpod,guan2025promptdino,fu2026petdino,qin2026vitprompt}. Unlike these representation-level approaches, \pqa delays explicit cross-modal state interaction until modality-specific candidate states have been formed.

Related ideas also appear in later-stage evidence fusion outside this exact setting. F-VLM fuses region scores from different models~\citep{kuo2023fvlm}, DeepInteraction preserves modality-specific representations while allowing interaction~\citep{yang2022deepinteraction}, and MS-DETR models sensor contributions through loosely coupled fusion~\citep{xing2024msdetr}.

\paragraph{Set prediction.}
DETR formulates detection as set prediction with permutation-invariant matching~\citep{carion2020detr}, and DINO further develops this formulation through denoising and improved anchor boxes~\citep{zhang2023dino}. Relation Networks and Learning NMS show that relations among candidate hypotheses can affect final decisions~\citep{hu2018relation,hosang2017nms}. \pqa combines these ideas at the candidate-set level: it operates on two candidate sets, establishes pairing through shared initialization rather than an explicit matching step, and applies a permutation-equivariant module to reason over the fused set.

\section{Method}

\subsection{Open-Prompt Detection as a Set-to-Set Mapping}

For a query image \(I\) and task category set \(\mathcal C\), each category \(c\) is specified by a text description \(t_c\), a support set of \(K\) boxed exemplars
\[
\mathcal S_c=\{(I_c^k,b_c^k)\}_{k=1}^{K},
\]
or both. We treat \(\mathcal S_c\) as a set: its order carries no information, while \(K\) is chosen by the user. The detector maps these prompts to a detection set through three set-valued stages:
\begin{equation}
\begin{aligned}
V_c &= \Phi_{\mathrm{BR}}(\mathcal S_c), \qquad |V_c|=R+1 \quad \forall K,\\
\{s_i^T\}_{i=1}^{N} &= \Psi_{T}\!\left(I,\{t_c\}_{c\in\mathcal C},Q^0\right),\qquad
\{s_i^V\}_{i=1}^{N} = \Psi_{V}\!\left(I,\{V_c\}_{c\in\mathcal C},Q^0\right),\\
Q^F &= \Psi_{\mathrm{PQA}}\!\left(\{(s_i^T,s_i^V)\}_{i=1}^{N}\right),\qquad
\hat{\mathcal Y}=\Head(Q^F).
\end{aligned}
\label{eq:setmap}
\end{equation}
Here \(\Phi_{\mathrm{BR}}\) is the \br support-set encoder (Sec.~\ref{sec:br}); the text reader \(\Psi_T\) and the visual reader \(\Psi_V\) decode the same image from a shared initialization \(Q^0\) of \(N=900\) queries, conditioned respectively on the text descriptions \(\{t_c\}_{c\in\mathcal C}\) and the \br memories \(\{V_c\}_{c\in\mathcal C}\); \(s_i^T\) and \(s_i^V\) are the detection states produced from the \(i\)-th initial query; and \(\Psi_{\mathrm{PQA}}\) fuses paired states into the set \(Q^F\), from which the prediction head outputs at most 300 detections (Sec.~\ref{sec:pqa}).

\begin{figure}[t]
\centering
\includegraphics[width=\textwidth]{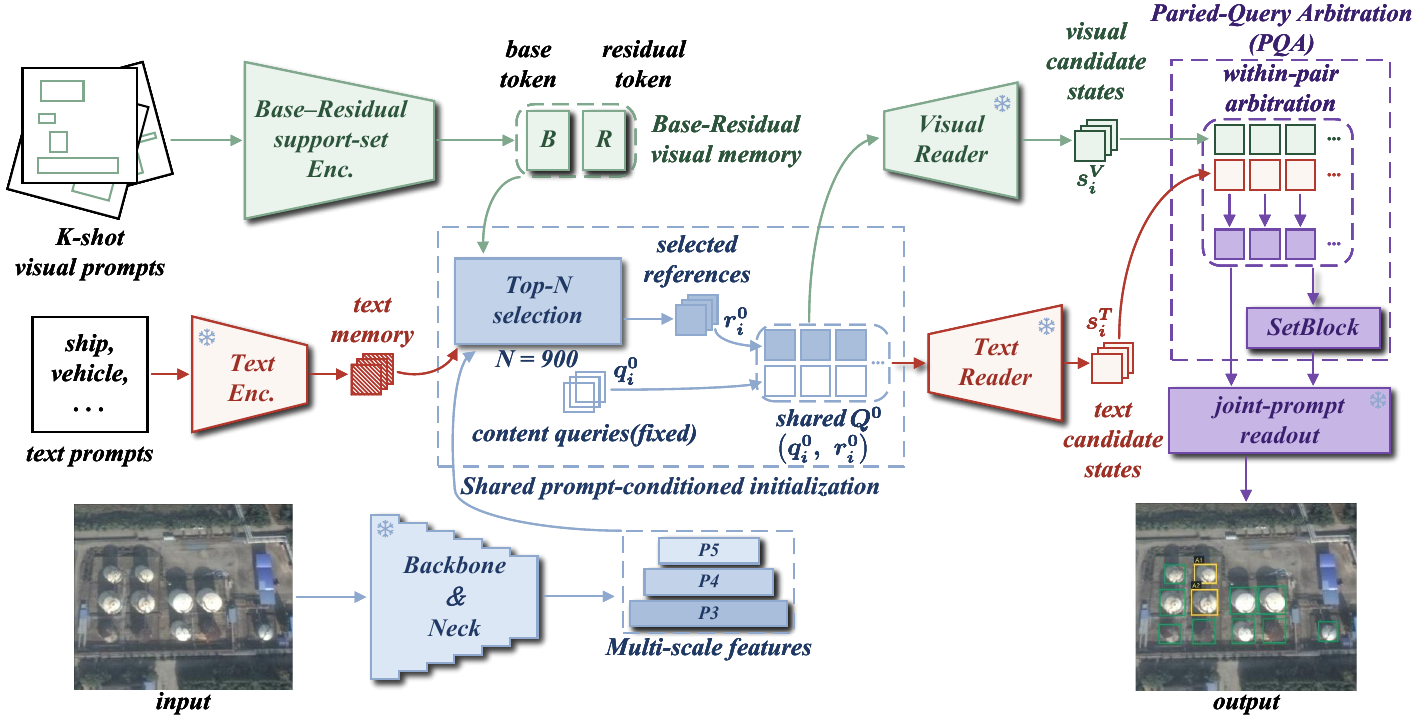}
\caption{\textbf{\setopd overview.} A support set of \(K\) boxed exemplars is encoded by the box-conditioned \br encoder into a fixed-capacity visual memory. Text and visual readers decode from a shared prompt-conditioned initialization, so their candidates are paired by index; \pqa arbitrates within each pair, reasons over the fused candidate set, and the frozen prompt-conditioned head produces the final detections. The visual memory also supports the native Visual-only pathway.}
\label{fig:overview}
\end{figure}

Figure~\ref{fig:overview} shows the overall architecture of \setopd. This set view yields three requirements, verified in Appendix~\ref{app:proofs}: (a) \(\Phi_{\mathrm{BR}}\) should be invariant to exemplar order and its output size should not depend on \(K\); (b) correspondence between the two candidate sets should not depend on query order; and (c) \(\Psi_{\mathrm{PQA}}\) should be permutation equivariant, so the model output is a set and the Hungarian set loss is independent of query order.

\subsection{Base--Residual Support-Set Encoding}
\label{sec:br}

\paragraph{Box-conditioned instance encoding.}
For each boxed support instance, a frozen prompt-independent image encoder extracts full-image multi-scale features \(F_{ck}^{P3:P5}\). Beyond appearance, the box conveys target position, scale, and shape through the six-dimensional geometry descriptor
\[
m_{ck}=[c_x,c_y,w,h,\log(w/h),\log(wh/(WH))],
\]
where \(W\) and \(H\) are the image width and height. The first four normalized coordinates serve as reference points for multi-scale deformable attention~\citep{zhu2021deformable}, which reads instance evidence from the full-image features; because each instance is read in its complete scene context rather than from an independently encoded crop, the readouts are context aware:
\begin{equation}
\begin{aligned}
q_{ck}^{\mathrm{box}}
&=\LN(\MLP_{\mathrm{box}}(m_{ck})),\\
e_{ck}^{(\ell,g)}
&=\LN\!\left(
W_e\,\operatorname{MSDeformAttn}^{\ell,g}
(q_{\mathrm{content}},q_{ck}^{\mathrm{box}},
F_{ck}^{P3:P5},m_{ck,1:4})
\right),
\end{aligned}
\label{eq:evidence}
\end{equation}
with \(\ell=1,\ldots,L\) feature levels (\(L=3\) for P3--P5) and \(g=1,\ldots,G\) sampling points per level. Instead of collapsing the \(L\!\cdot\!G\) sampled responses of an exemplar into a single token, the encoder retains them, after a shared projection \(W_e\) and normalization, as evidence vectors; the support set thus produces \(E_c=\{e_{cn}\}_{n=1}^{N_e}\) with \(N_e=KLG\), padded positions masked out.

\paragraph{Base--Residual support-set representation.}
The Base token summarizes the evidence set, and each Residual slot pools deviations from the Base:
\begin{equation}
\mu_c=\frac{1}{N_e}\sum_{n=1}^{N_e}e_{cn},
\qquad
B_c=\mathcal B(\mu_c),
\qquad
R_{cr}=\mathcal R_r
\!\left(\{e_{cn}-\sg(B_c)\}_{n=1}^{N_e}\right),
\label{eq:br}
\end{equation}
where \(\mathcal B\) is a single-seed mean encoder~\citep{lee2019settransformer,zaheer2017deepsets} and \(\mathcal R_r\) is a slot update that follows the normalization pattern of Slot Attention~\citep{locatello2020slot} but pools deviations from a support-dependent anchor with deterministic slot queries and residual feed-forward steps rather than a GRU; \(\sg(\cdot)\) denotes stop-gradient, and full parameterizations are given in Appendix~\ref{app:param}. The visual memory is
\begin{equation}
V_c=[B_c,R_{c1},\ldots,R_{cR}],\qquad |V_c|=R+1,
\label{eq:visualmemory}
\end{equation}
so the prompt size is always \(R+1\), regardless of \(K\). Our default is \(R=1\), giving two tokens (T2). T1, T3, and T4 use \(R=0,2,3\), respectively, and serve as structural controls in Table~\ref{tab:br}B.

\paragraph{Readout and the Visual-only pathway.}
For candidate query \(z_i\), the visual memory yields a category score through gated Residual responses,
\begin{equation}
s_{ic}^{V}
=
z_i^\top B_c
+
\lambda_R
\frac{
\sum_r g_{icr}\,z_i^\top R_{cr}
}{
\sum_r g_{icr}+\epsilon
},
\qquad
g_{icr}
=
\sigma\!\left(
\frac{z_i^\top R_{cr}}{\sqrt d\,\tau}
\right),
\label{eq:visualreadout}
\end{equation}
with \(\lambda_R=0.5\) and \(\tau=1.0\); for the default \(R=1\), this reduces to the fixed combination \(z_i^\top(B_c+\lambda_R R_{c1})\). The structured visual memory plugs directly into the detector's native prompt-conditioned pathway and drives category-dependent candidate scoring, selection of the top 900 queries, decoder--prompt interaction, and final classification, forming a complete Visual-only detector.

\subsection{Paired-Query Arbitration over Candidate Sets}
\label{sec:pqa}

\paragraph{Shared prompt-conditioned initialization.}
Text and visual prompts are jointly used only to select the top-\(N\) proposal references, while the learnable content queries remain fixed. Both readers then decode from the same initialization \(Q^0=\{(q_i^0,r_i^0)\}_{i=1}^{N}\) against their own prompt memories, producing modality-specific states \(s_i^m=(h_i^m,r_i^m,p_i^m,b_i^m)\) (hidden state, reference, class probabilities, and box), \(m\in\{T,V\}\). Hence, same-index text and visual candidates form natural initialization pairs without explicit cross-modal matching. This pairing is a structural prior rather than a hard object-level correspondence; an audit in Appendix~\ref{app:pairing} shows that it is nevertheless highly consistent for valid detections.

\paragraph{Within-pair arbitration.}
For pair \(i\), the arbitrator receives both hidden states, their absolute difference, the maximum class confidence from each branch, and the difference between the two references in inverse-sigmoid space---a 774-dimensional descriptor (\(3\times256+2+4\), with hidden dimension \(d=256\))---and produces a scalar text weight \(g_i\) that modulates both the semantic state and the geometric reference:
\begin{equation}
\begin{aligned}
x_i &=
[h_i^T,h_i^V,|h_i^T-h_i^V|,
\max_c p_i^T(c),\max_c p_i^V(c),
\logitfn(r_i^T)-\logitfn(r_i^V)],\\
g_i &= \sigma(\MLP(x_i)), \qquad \MLP: 774\!\rightarrow\!128\!\rightarrow\!1,\\
h_i^P &= g_i h_i^T+(1-g_i)h_i^V,\qquad
r_i^P =
\sigma\!\left(
g_i\logitfn(r_i^T)+(1-g_i)\logitfn(r_i^V)
\right).
\end{aligned}
\label{eq:pqa}
\end{equation}
A constant gate recovers fixed rules as special cases: text only (\(g_i=1\)), visual only (\(g_i=0\)), and mean fusion (\(g_i=1/2\)); a saturated gate selects one candidate from each pair. The learned gate chooses among these behaviors according to the evidence in \(x_i\).

\paragraph{From pairs to set-level reasoning.}
Pairwise arbitration processes each pair independently and therefore cannot account for other candidates, for example when two pairs describe the same object. We therefore treat the \(N\) fused states as a set~\citep{carion2020detr,zhang2023dino} and apply one set-reasoning block,
\begin{equation}
H^F=
\operatorname{SetBlock}(H^P),
\qquad
H^P=[h_1^P,\ldots,h_N^P],
\label{eq:setreasoning}
\end{equation}
where \(\operatorname{SetBlock}\) is one eight-head self-attention layer without index-dependent positional encoding followed by a \(256\!\rightarrow\!512\!\rightarrow\!256\) FFN, both with residual connections and LayerNorm (full form in Appendix~\ref{app:param}); the references remain \(r_i^P\).

\paragraph{Frozen joint-prompt readout.}
The final classifier does not combine the original unimodal reader scores. Instead, both modality-conditioned probabilities are recomputed from the same fused state \(h_i^F\); box offsets are decoded from \((h_i^F,r_i^P)\) by the frozen regression branch, where \(R^P=[r_1^P,\ldots,r_N^P]\):
\begin{equation}
\begin{aligned}
p_{ic}^{T,F}
&=
\frac{1}{|\mathcal T_c|}
\sum_{j\in\mathcal T_c}
\sigma\!\left(\ell^T(h_i^F,t_j)\right),\qquad
p_{ic}^{V,F}
=
\sigma\!\left(\ell^V(h_i^F,V_c)\right),\\
s_{ic}^{F}
&=
\logitfn\!\left[
\operatorname{clip}\!\left(
\tfrac12 p_{ic}^{T,F}
+\tfrac12 p_{ic}^{V,F},
\;10^{-7},\,1-10^{-7}
\right)\right],
\end{aligned}
\label{eq:readout}
\end{equation}
where \(\ell^T\) is the frozen text-conditioned classifier averaged over the text tokens \(\mathcal T_c\) of category \(c\), and \(\ell^V\) is the frozen \br readout of Eq.~(\ref{eq:visualreadout}). The final \(1{:}1\) fusion is performed in probability space on this single fused state, and category semantics remain carried entirely by the prompt tokens.

\paragraph{Structural properties.}
The \br encoder is permutation invariant to exemplar order and has a fixed output size of \(R+1\) tokens for any \(K\). With the default \(R=1\), it implements a mean-dependent anchor--correction factorization rather than instance-level routing. \pqa is permutation equivariant with respect to the shared query ordering, and pairing by initialization costs \(O(N)\) instead of the \(O(N^3)\) assignment that aligning two independently decoded candidate sets would require. Formal statements and proofs are given in Appendix~\ref{app:proofs}.

\subsection{Learning Objective}

\paragraph{Training objective.}
We use the standard DETR Hungarian detection loss. In addition, for each matched fused query, we penalize its detection error when it exceeds that of the better paired unimodal reader:
\[
\mathcal L
=
\mathcal L_{\mathrm{det}}
+0.1\,\operatorname{mean}_i
\max\!\left(0,\,e_i^F-\min(e_i^T,e_i^V)\right).
\]
The unimodal errors are stop-gradient targets, so this term constrains only the fused state; full loss definitions and the staged training protocol, which concludes with a quality-focal-loss continuation producing the final model, are given in Appendix~\ref{app:training}.

\section{Experiments}

\subsection{Experimental Setup}

\paragraph{Data.}
We construct an 11-source remote-sensing OPD evaluation benchmark under a unified 153-category semantic space. The training corpus and category harmonization follow the multi-source construction introduced in prior work~\citep{hu2026hiopd} (175{,}644 training images over 153 categories, integrating sources with different taxonomies, annotation granularities, and scene distributions), while the evaluation protocol in this paper uses seven in-domain test sources---DIOR, DOTA-v2.0, FAIR1M, SIMD, SODA-A, ShipRSImageNet, and WHU-Buildings---and four fully held-out sources that are never used during training: HRSC2016, HRRSD, RSOD, and VEDAI. We therefore report Macro11 over seven in-domain test sources and four held-out sources. Main experiments report COCO-style AP, and comparisons with external methods also report AP50 when available. Support exemplars and query images always come from non-overlapping scenes. The experiments address two questions: how a few visual exemplars can form an effective visual prompt for a frozen detector (Sec.~\ref{sec:exp_br}), and whether, beyond joint prompting, modality-specific candidate states expose complementary detection evidence that can be explicitly arbitrated (Sec.~\ref{sec:exp_pqa}).

\paragraph{Implementation details.}
Training has three stages: Stage~0 adapts Grounding DINO~\citep{liu2024groundingdino} with a Swin-T backbone and BERT text encoder on the RS153 corpus into the prompt-neutral base endpoint, which is frozen thereafter; Stage~I trains only the \br visual-prompt encoder; and Stage~II trains only the \pqa pair gate and one-layer set adapter. We use \(K=5\) boxed exemplars per category. Full optimization details---iterations, batch sizes, optimizer, seeds, and the final quality-focal-loss continuation that produces the reported model---are given in Appendix~\ref{app:training}.

\subsection{Comparison with Open-Prompt Detectors}

We compare \setopd with existing open-prompt detectors on DIOR and DOTA-v2.0. Because published methods use different training data and evaluation protocols, the complete reported-results comparison, organized by text, visual, and text+visual prompting modes, is provided in Appendix~\ref{app:external} (Table~\ref{tab:benchmark}). Within \setopd, text+visual prompting raises Macro11 AP over both unimodal readers (55.57 vs.\ 53.92 and 53.55): on DIOR it improves over both (57.7/58.2 \(\to\) 60.0), and on DOTA-v2.0 it improves over the visual reader (46.0 \(\to\) 49.0) while remaining on par with the strong text reader (49.4). In the reported values, \setopd also obtains the highest DOTA-v2.0 AP among the methods for which AP is available.

\subsection{Ablation and Analysis of Visual Prompt Encoding}
\label{sec:exp_br}

Table~\ref{tab:br} contains two complementary studies: Part~A analyzes sensitivity to the number of support exemplars, while Part~B performs structural ablations of the \br representation; the equal-token mean-prototype control matches the prompt length of Base+R1 without the anchor--correction decomposition.

\begin{table*}[t]
\centering
\caption{Ablation and analysis of \br visual prompting. Part~A studies support-set size; Part~B isolates the contributions of Base, Residual, prompt length, and multi-Residual routing. All structural variants are independently trained from the same frozen Stage-0 endpoint with the same optimization budget and support-sampling protocol. All values are Visual-only AP.}
\label{tab:br}
\scriptsize
\setlength{\tabcolsep}{2.2pt}
\begin{adjustbox}{max width=\textwidth}
\begin{tabular}{lrrrrrrrrrrrr}
\toprule
Setting & DIOR & DOTA & FAIR1M & HRSC & SIMD & SODA-A & Ship & WHU & HRRSD & RSOD & VEDAI & Macro11 \\
\midrule
\multicolumn{13}{c}{\textbf{A. Support-example efficiency}}\\
\midrule
Mean prototype, \(K=1\) & 55.4 & 42.6 & 23.5 & 63.3 & 59.5 & 37.9 & 47.3 & 51.3 & 26.7 & 0.4* & 55.3 & 42.11 \\
\br, \(K=1\) & 56.0 & 41.3 & 25.9 & 70.5 & 68.9 & 39.0 & 48.0 & 78.9 & 30.1 & 42.7 & 53.7 & 50.45 \\
\br, \(K=3\) & 57.3 & 45.4 & 27.5 & 74.7 & 70.2 & 39.5 & 57.0 & 78.7 & 31.3 & 40.2 & 51.1 & 52.08 \\
\br, \(K=5\) & 58.2 & 46.0 & 28.8 & 78.4 & 71.6 & 39.5 & 58.2 & 78.9 & 32.4 & 42.7 & 54.4 & 53.55 \\
\br, \(K=10\) & 58.4 & 48.3 & 28.5 & 73.5 & 71.6 & 39.4 & 58.9 & 79.2 & 32.1 & 43.0 & 56.4 & 53.57 \\
\midrule
\multicolumn{13}{c}{\textbf{B. Structural attribution (\(K=5\))}}\\
\midrule
Base+R1 & 58.2 & 46.0 & 28.8 & 78.4 & 71.6 & 39.5 & 58.2 & 78.9 & 32.4 & 42.7 & 54.4 & 53.55 \\
Residual only & 36.0 & 25.9 & 13.3 & 41.4 & 43.1 & 33.2 & 17.8 & 27.2 & 21.7 & 36.8 & 31.0 & 29.76 \\
Mean prototype \(\times2\) & 58.1 & 47.6 & 27.7 & 74.2 & 68.6 & 43.7 & 57.3 & 66.8 & 28.3 & 0.44* & 51.73 & 47.68 \\
Base only & 57.8 & 47.1 & 27.5 & 77.8 & 70.6 & 39.0 & 56.7 & 63.5 & 28.8 & 0.1* & 49.3 & 47.11 \\
Base+2R (T3) & 58.1 & 46.3 & 28.4 & 79.8 & 71.1 & 39.3 & 59.7 & 78.8 & 31.9 & 42.2 & 53.4 & 53.55 \\
Base+3R (T4) & 58.2 & 46.4 & 28.8 & 79.1 & 71.5 & 39.6 & 58.1 & 79.0 & 32.1 & 42.4 & 52.2 & 53.40 \\
\bottomrule
\end{tabular}
\end{adjustbox}
\vspace{2pt}
\par\smallskip{\scriptsize *Entries with severe performance degradation.}\par
\end{table*}

Three conclusions suffice. First, a single exemplar already reaches 94\% of the \(K=5\) Macro11 AP (50.45 vs.\ 53.55), and performance saturates around \(K=5\) while the prompt length stays fixed at two tokens. Second, the anchor--correction factorization is substantially more effective than directly using or duplicating the pooled prototype in our implementation: Base alone reaches 47.11 AP, adding one Residual correction raises it to 53.55 (\(+6.4\) AP), and the equal-token mean-prototype control reaches only 47.68 AP, so prompt length alone does not explain the gap. Third, multi-Residual routing does not help: Base+2R and Base+3R score no higher than Base+R1. The mean-prototype and Base-only variants suffer severe performance degradation on the held-out RSOD source, where Base+R1 restores detection; even excluding RSOD, Base+R1 still improves over Base-only by 2.83 AP on average across the remaining ten sources. This suggests that the anchor--correction parameterization improves both average accuracy and robustness across source domains.

\subsection{Candidate-State Collaboration Outperforms Representation-Level Fusion}
\label{sec:exp_pqa}

To compare candidate-set-level late fusion with common representation-level fusion and candidate-relation strategies, we report two unimodal reference endpoints followed by collaboration baselines within the same base detection framework, with every collaboration baseline built on the same \br visual reader. The text reader is the frozen Stage-0 text pathway of the adapted Grounding DINO~\citep{liu2024groundingdino} base detector, and the \br visual reader is our own visual branch (Sec.~\ref{sec:br}), decoded under the shared query initialization of the joint pathway rather than run as the native Visual-only detector of Table~\ref{tab:br}---its per-source APs therefore differ slightly from the Visual-only rows (e.g., 47.30 vs.\ 46.0 on DOTA-v2.0). Joint Prompt Representation~\citep{yang2026rsmpod,xu2023mqdet} fuses text and vision at the prompt/representation level, and Relation-Aware Selection~\citep{hu2018relation,hosang2017nms} models object relations at the candidate level. Three further adaptations---Sparse Query Fusion~\citep{carion2020detr}, MS-DETR~\citep{xing2024msdetr}, and DeepInteraction~\citep{yang2022deepinteraction}---were adapted on only four shared sources and are reported in Appendix~\ref{app:adapted}.

\begin{table*}[t]
\centering
\caption{Comparison of unimodal readers and text--visual collaboration strategies on all 11 sources (AP).}
\label{tab:collab}
\scriptsize
\setlength{\tabcolsep}{2.0pt}
\begin{adjustbox}{max width=\textwidth}
\begin{tabular}{lrrrrrrrrrrrr}
\toprule
Method & DIOR & DOTA & FAIR & HRSC & SIMD & SODA & ShipRS & WHU & HRRSD & RSOD & VEDAI & M11 \\
\midrule
Text Reader (Stage-0, frozen) & 57.75 & \textbf{49.45} & 29.78 & 78.10 & 72.08 & 39.36 & 60.38 & 79.41 & 33.05 & 40.49 & 53.22 & 53.92 \\
\br Visual Reader (Ours) & 58.11 & 47.30 & 28.99 & 78.84 & 71.82 & 39.61 & 58.89 & 79.02 & 31.35 & 41.67 & 53.46 & 53.55 \\
Joint Prompt Representation & 58.90 & 48.93 & 29.25 & 79.24 & 71.98 & 39.66 & 60.68 & 79.27 & 32.16 & 39.96 & 54.00 & 54.00 \\
Relation-Aware Selection & 58.35 & 46.71 & 29.90 & 79.18 & 71.18 & 39.80 & 61.90 & 78.51 & \textbf{33.44} & 41.17 & 54.91 & 54.10 \\
\midrule
\setopd (Ours) & \textbf{59.97} & 49.01 & \textbf{30.54} & \textbf{79.89} & \textbf{72.54} & \textbf{44.59} & \textbf{62.66} & \textbf{83.06} & 32.19 & \textbf{41.91} & \textbf{54.93} & \textbf{55.57} \\
\bottomrule
\end{tabular}
\end{adjustbox}
\end{table*}

Table~\ref{tab:collab} first reports the two unimodal readers as reference endpoints, followed by three text--visual collaboration strategies. \setopd reaches 55.57 Macro11 AP across all 11 sources, improving over the text reader (53.92), \br visual reader (53.55), Joint Prompt Representation (54.00), and Relation-Aware Selection (54.10) by 1.66, 2.02, 1.57, and 1.48 AP, respectively. At the source level, \setopd obtains the highest AP in the table on nine of 11 sources, including the held-out RSOD and VEDAI, indicating that the overall improvement is not driven by a single source. Beyond its average gain, \setopd also exceeds the \emph{stronger} of the two unimodal readers on 9 of 11 sources and remains within 1 AP of it on the remaining two (DOTA-v2.0 and HRRSD), indicating that candidate-state collaboration often produces gains beyond simply selecting the better modality.

More importantly, relative to representation-level Joint Prompt Representation, \pqa achieves higher AP on all 11 sources and raises Macro11 from 54.00 to 55.57. The gain is not confined to in-domain sources: \pqa improves over Joint Prompt Representation by 1.96 AP on the seven in-domain sources and by 0.89 AP on the four fully held-out sources. RS-MPOD's cross-dataset and fine-grained experiments likewise show that representation-level multimodal prompting can underperform visual prompting on some external datasets, which its authors attribute to semantic noise from mismatched text cues during fusion~\citep{yang2026rsmpod}. Because the training data and evaluation protocols differ, we do not make direct numerical comparisons to those results. Our controlled comparison within a shared base framework instead shows that delaying explicit cross-modal state fusion until text and vision have independently formed candidate detections more consistently preserves useful evidence from both modalities.

\paragraph{Diagnosing the DOTA-v2.0 sensitivity.}
DOTA-v2.0 exposes a residual sensitivity in the final scoring stage. Simply continuing \pqa training brings only a marginal gain, whereas using the frozen Text-reader boxes recovers noticeably more AP. Removing the best-reader preservation term has almost no additional effect, while quality-aware classification provides a further improvement. More importantly, a zero-training counterfactual that retains the \pqa candidate list but replaces its scores with Text-reader scores recovers substantially more AP. This indicates that the residual DOTA-v2.0 degradation is dominated by score/query-class ranking and its coupling with localization, rather than by the absence of useful multimodal candidate states. A diagnostic ablation further localizes the sensitivity to the final score/readout stage; zero-training readout probes provide additional evidence that useful multimodal candidates are already present (Appendix~\ref{app:dota_diagnosis}).

Detection-level analysis further confirms this behavior: both readers contribute modality-exclusive correct detections, while \pqa retains 53.8--100\% of their union across four audited sources (Appendix~\ref{app:evidence}).

\section{Conclusion}

\setopd shows that multimodal open-prompt detection need not collapse text and visual evidence in representation space. Across 11 heterogeneous remote-sensing sources, preserving modality-specific candidate states and reasoning over them as a set consistently improves over representation-level joint prompting. Together with the observed detection-level complementarity, these results support candidate-set reasoning as an alternative design principle for multimodal prompt integration.

\subsection*{AI use statement}

Generative AI tools were used to assist with language editing, including improving the clarity, concision, and organization of the manuscript; literature search and organization; and code implementation and debugging. All AI-assisted content and code were reviewed and verified by the authors, who take full responsibility for the final manuscript and implementation.

\subsection*{Reproducibility statement}

Section~3 and Appendices~\ref{app:training}--\ref{app:param} specify the model architecture, layer dimensions, training objectives, hyperparameters, and the staged optimization protocol, including \(N=900\), \(R=1\), \(\lambda_R=0.5\), \(\tau=1.0\), detection-loss weights \(1/5/2\), the \(\mathcal L_{\mathrm{keep}}\) weight \(0.1\), iteration counts, batch sizes, optimizer, and seeds. Appendix~\ref{app:proofs} proves Propositions~1 and 2. The supplementary material additionally provides anonymized implementation notes and module-level pseudocode documenting the interfaces and tensor flows of the proposed components; the complete source code and configuration files will be released upon acceptance.

\FloatBarrier
\bibliographystyle{iclr2027_conference}
\bibliography{references}

\clearpage
\appendix

\section{Supplementary Experiments and Details}

This appendix reports diagnostic analyses of \pqa, detection-evidence complementarity, the initialization-pairing audit, comparisons with external open-prompt detectors, adapted collaboration baselines, and full training-objective and implementation details.

\subsection{Diagnostic Ablation and Readout Analysis on DOTA-v2.0}
\label{app:dota_diagnosis}

We further diagnose why candidate-state collaboration is more sensitive on DOTA-v2.0. The analysis separates four possible factors: insufficient optimization, the localization path used by the fused candidates, the best-reader preservation loss, and classification-score calibration. All trained controls start from the same \pqa checkpoint and use the same additional training budget and optimization protocol. We additionally include two zero-training readout counterfactuals to test whether useful candidate states already exist before score repair.

Table~\ref{tab:dota_diagnosis} shows that additional optimization alone changes AP only marginally. Replacing the fused localization path with the frozen Text-reader boxes gives a larger recovery, whereas removing the best-reader preservation loss provides almost no further gain. With the box policy and preservation setting fixed, QFL classification adds another 0.30 AP. The strongest evidence comes from the zero-training controls: retaining the \pqa candidate list while replacing its scores with Text-reader scores reaches 50.14 AP, and logit averaging with Text-reader boxes reaches 50.17 AP. Thus, the DOTA-v2.0 gap is primarily associated with score/query-class ranking and score--localization coupling rather than failure to form useful multimodal candidates.

\begin{table}[H]
\centering
\caption{
Diagnostic ablation and zero-training readout analysis on DOTA-v2.0.
The first group isolates optimization, localization policy,
the preservation loss, and QFL classification;
the second group probes the final scoring/readout without additional training.
}
\label{tab:dota_diagnosis}
\small
\setlength{\tabcolsep}{5.5pt}
\begin{tabular}{lcc}
\toprule
Setting & DOTA AP & \(\Delta\) AP \\
\midrule
\multicolumn{3}{c}{\textit{Diagnostic ablations}} \\
\midrule
PQA checkpoint & 48.25 & -- \\
$+$ continued optimization & 48.33 & $+$0.08 \\
$+$ Text-reader localization & 48.67 & $+$0.34 \\
$-$ preservation loss & 48.71 & $+$0.04 \\
$+$ QFL classification (Final \setopd) & 49.01 & $+$0.30 \\
\midrule
\multicolumn{3}{c}{\textit{Zero-training readout probes}} \\
\midrule
PQA candidate list + Text-reader scores & 50.14 & $+$1.89$^{\dagger}$ \\
Text/PQA logit averaging + Text-reader boxes & 50.17 & $+$1.92$^{\dagger}$ \\
\bottomrule
\end{tabular}
\vspace{2pt}
\par\smallskip{\scriptsize
For the first five rows, \(\Delta\) AP is relative to the immediately preceding matched control.
$^{\dagger}$Zero-training counterfactual; \(\Delta\) is relative to the PQA checkpoint.
}\par
\end{table}

\subsection{Detection-Evidence Complementarity}
\label{app:evidence}

At IoU threshold 0.5, we partition ground-truth objects into those detected only by the text reader (Text-only GT), only by the visual reader (Visual-only GT), and by both readers (Common GT). The modality-exclusive set is the union of the two exclusive subsets, denoted Exclusive GTs, and \pqa retention is the fraction of these exclusive ground-truth objects that \pqa still detects. Counting unique ground-truth objects rather than raw detections ensures that duplicate detections of the same object are not counted multiple times. This analysis directly quantifies complementarity at the detection level.

\begin{table}[H]
\centering
\caption{Detection-level complementarity between the text and visual readers at IoU 0.5, counted over unique ground-truth objects. Exclusive GTs is the union of Text-only and Visual-only ground-truth objects; \pqa retained counts how many of them \pqa still detects.}
\label{tab:evidence}
\scriptsize
\setlength{\tabcolsep}{3pt}
\begin{tabular}{lrrrrr}
\toprule
Dataset & Text-only GT & Visual-only GT & Common GT & Exclusive GTs & \pqa retained \\
\midrule
DIOR & 698 & 1,716 & 116,758 & 2,414 & 1,763 / 2,414 (73.03\%) \\
DOTA-v2.0 & 1,876 & 2,196 & 134,463 & 4,072 & 2,191 / 4,072 (53.81\%) \\
HRSC2016 & 1 & 18 & 1,209 & 19 & 19 / 19 (100.00\%) \\
RSOD & 12 & 71 & 7,103 & 83 & 70 / 83 (84.34\%) \\
\bottomrule
\end{tabular}
\end{table}

\begin{figure}[H]
    \centering
    \includegraphics[width=\textwidth]{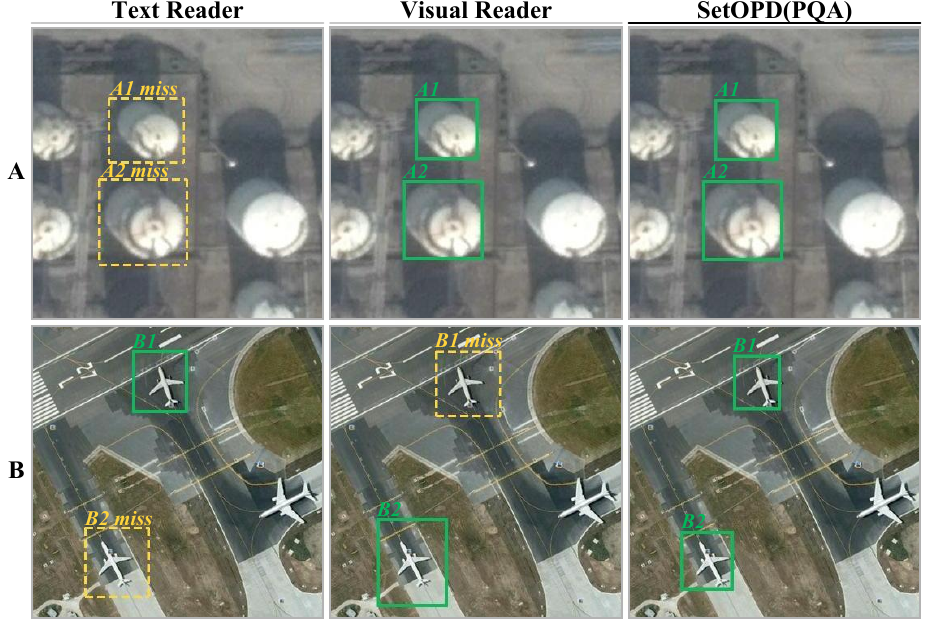}
    \caption{\textbf{Qualitative detection-level complementarity and \pqa arbitration.}
    Columns show ground truth and the predictions of each reader.
    Green boxes denote matched predictions, while yellow dashed boxes indicate
    selected ground-truth objects missed by the corresponding reader.
    Top: the visual reader recovers storage tanks missed by the text reader,
    and \pqa preserves them.
    Bottom: the two readers recover complementary airplane instances,
    while \pqa retains both.
    All columns within each row use the same image crop.}
    \label{fig:qualitative}
\end{figure}

Both readers contribute modality-exclusive ground-truth objects on all four sources, and \pqa retains them unevenly: from 100.00\% on HRSC2016 down to 73.03\% on DIOR and 53.81\% on DOTA-v2.0. Retention below 100\% quantifies how much single-reader evidence is lost when disagreeing pairs are resolved into a single decision. Notably, the visual reader contributes more modality-exclusive ground-truth objects than the text reader on all four audited sources, further indicating that visual prompting supplies non-redundant instance-level evidence. Figure~\ref{fig:qualitative} shows qualitative examples: the visual reader recovers storage tanks missed by the text reader, and in another image the two readers detect complementary airplane instances; \pqa retains both.

\subsection{Initialization-Pairing Audit}
\label{app:pairing}

To quantify how reliably shared initialization induces object-level correspondence, we audit the text reader and the \br visual reader on three sources (DIOR, DOTA-v2.0, and HRSC2016; 128 images each, fixed sampling). Table~\ref{tab:pairing} reports three quantities. First, conditioned on both branches producing class-correct true positives (IoU \(\ge 0.5\)), same-index pairs refer to the same ground-truth object in at least 99.86\% of the qualifying pairs on each audited source. Second, same-index reference boxes are spatially well aligned (mean IoU 0.767--0.857). Third, compared with a box-IoU-optimal one-to-one Hungarian reassignment within the Top-50 selected queries, the original index pairing is retained for 83.94--94.03\% of indices.

Same-GT is a conditional measure: it should not be interpreted as correspondence accuracy over all selected queries, since most queries do not satisfy the strict verification condition that both branches be class-correct true positives. Likewise, index pairing agrees with box-IoU-optimal assignment for most Top-50 candidates, while the remaining 5.97--16.06\% shows that shared initialization is not a universal correspondence guarantee. Shared initialization therefore provides a strong index-wise correspondence prior for valid detections, and the residual disagreement is one reason \pqa performs within-pair arbitration and set-level reasoning instead of averaging same-index states directly.

\begin{table}[H]
\centering
\caption{Initialization-pairing audit over three sources (128 images each).}
\label{tab:pairing}
\scriptsize
\setlength{\tabcolsep}{4pt}
\begin{tabular}{lccc}
\toprule
Dataset & Same-GT given both TP & Same-index box IoU & Index retained after Hungarian \\
\midrule
DIOR & 99.90\% & 0.854 & 94.03\% \\
DOTA-v2.0 & 99.86\% & 0.857 & 93.97\% \\
HRSC2016 & 100.00\% & 0.767 & 83.94\% \\
\bottomrule
\end{tabular}
\vspace{2pt}
\par\smallskip{\scriptsize Same-GT is conditioned on both branches being class-correct TPs at IoU \(\ge 0.5\). It should not be interpreted as correspondence accuracy over all selected queries.}\par
\end{table}

\subsection{Reported Results of External Open-Prompt Detectors}
\label{app:external}

Table~\ref{tab:benchmark} compares \setopd with existing open-prompt detectors on DIOR and DOTA-v2.0 using results reported in their source papers.

\begin{table}[H]
\centering
\caption{Reported-results comparison with existing open-prompt detectors on DIOR and DOTA-v2.0. Dashes indicate metrics not reported in the source papers.}
\label{tab:benchmark}
\scriptsize
\setlength{\tabcolsep}{4pt}
\begin{tabular}{l c c c c}
\toprule
Method & DIOR AP & DIOR AP50 & DOTA-v2.0 AP & DOTA-v2.0 AP50 \\
\midrule
Grounding DINO~\citep{liu2024groundingdino} (Text) & -- & 78.7 & -- & 71.8 \\
LAE-DINO~\citep{pan2025laedino} (Text) & -- & 85.5 & 46.8 & -- \\
OpenRSD~\citep{huang2025openrsd} (Text) & -- & 76.7 & -- & 71.8 \\
OpenRSD~\citep{huang2025openrsd} (Visual) & -- & 76.7 & -- & 69.8 \\
Hi-OPD~\citep{hu2026hiopd} (Text) & -- & 79.7 & 48.2 & 72.3 \\
Hi-OPD~\citep{hu2026hiopd} (Visual) & -- & 75.2 & 46.1 & 69.5 \\
WeDetect-tiny~\citep{fu2026wedetect} & -- & 76.5 & 42.6 & 65.6 \\
YOLO-World-L~\citep{cheng2024yoloworld} & -- & 73.2 & -- & 58.0 \\
RS-MPOD~\citep{yang2026rsmpod} (Text) & -- & 81.3 & 46.1 & 73.8 \\
RS-MPOD~\citep{yang2026rsmpod} (Text+Visual) & -- & 83.6 & 48.0 & 74.6 \\
RS-MPOD~\citep{yang2026rsmpod} (Visual, 32 inst.) & -- & 76.2 & 44.1 & 70.0 \\
\midrule
\setopd (Text) & 57.7 & 80.3 & 49.4 & 73.6 \\
\setopd (Visual, 5 inst.) & 58.2 & 80.6 & 46.0 & 68.8 \\
\setopd (Text+Visual) & \textbf{60.0} & 81.7 & \textbf{49.0} & 71.6 \\
\bottomrule
\end{tabular}
\end{table}

\subsection{Adapted Baselines on Four Shared Sources}
\label{app:adapted}

Sparse Query Fusion is an adaptation built from the sparse-query set formulation of DETR~\citep{carion2020detr}, rather than a method name from that work; MS-DETR Adaptation~\citep{xing2024msdetr} and DeepInteraction Adaptation~\citep{yang2022deepinteraction} transfer, respectively, loosely coupled multispectral fusion and Camera--LiDAR two-stream interaction to text/visual prompting. They were adapted on only four shared sources (DIOR, DOTA-v2.0, HRSC2016, and RSOD), so Macro11 is not reported; \setopd outperforms all three on these sources (Table~\ref{tab:adapted}).

\begin{table}[H]
\centering
\caption{Adapted collaboration baselines on the four shared sources (AP).}
\label{tab:adapted}
\scriptsize
\setlength{\tabcolsep}{4pt}
\begin{tabular}{lcccc}
\toprule
Method & DIOR & DOTA & HRSC & RSOD \\
\midrule
Sparse Query Fusion~\citep{carion2020detr} & 18.85 & 11.59 & 42.79 & 13.14 \\
MS-DETR Adaptation~\citep{xing2024msdetr} & 27.94 & 25.37 & 30.92 & 37.62 \\
DeepInteraction Adaptation~\citep{yang2022deepinteraction} & 53.04 & 43.66 & 70.45 & 38.72 \\
\midrule
\setopd (Ours) & \textbf{59.97} & \textbf{49.01} & \textbf{79.89} & \textbf{41.91} \\
\bottomrule
\end{tabular}
\end{table}

\subsection{Training-Objective Details}
\label{app:training}

Training has three stages. Stage~0 adapts Grounding DINO~\citep{liu2024groundingdino} with a Swin-T visual backbone and BERT text encoder on the RS153 training corpus, producing the prompt-neutral base endpoint, which is frozen in all subsequent stages. We use \(K=5\) boxed exemplars per category to build the default Base+R1 two-token visual memory. Stage~I trains only the \br visual-prompt encoder for 10,000 iterations with effective batch size 16. Stage~II trains only the \pqa pair gate and one-layer set adapter for 1,200 iterations with effective batch size 4 (about 0.63M parameters). Both stages use AdamW with learning rate \(10^{-4}\) and weight decay \(10^{-4}\); random seeds are 0 and 71, respectively.

\pqa is trained with DETR-style Hungarian matching and the standard classification, \(\ell_1\), and GIoU losses, weighted by \(1\), \(5\), and \(2\):
\begin{equation}
\mathcal L_{\mathrm{det}}
=
\mathcal L_{\mathrm{cls}}
+5\mathcal L_{\mathrm{L1}}
+2\mathcal L_{\mathrm{GIoU}}.
\label{eq:detloss}
\end{equation}
For a fused query \(i\) matched to ground-truth object \(j\), we compare only the text, visual, and fused states that share the same index \(i\). The matched detection error of each path is
\begin{equation}
e_i^m
=
-\log p_i^m(c_j)
+5\|b_i^m-b_j\|_1
+2\bigl(1-\GIoU(b_i^m,b_j)\bigr),
\qquad m\in\{T,V,F\}.
\label{eq:patherror}
\end{equation}
We then use
\begin{align}
\mathcal L_{\mathrm{keep}}
&=
\operatorname{mean}_i
\operatorname{ReLU}\!\left(
e_i^F-\min(e_i^T,e_i^V)
\right), \label{eq:keeploss}\\
\mathcal L
&=
\mathcal L_{\mathrm{det}}
+0.1\mathcal L_{\mathrm{keep}}. \label{eq:totalloss}
\end{align}
The unimodal errors \(e_i^T\) and \(e_i^V\) are treated as constants through stop-gradient, so \(\mathcal L_{\mathrm{keep}}\) constrains only the fused state. A fused query is penalized when it fits its matched target worse than the better of its two paired readers. The term encourages \pqa to preserve stronger unimodal evidence, but does not guarantee a higher final AP. After this objective converges, training concludes with the fused candidates localized by frozen text-reader boxes and a quality-focal-loss (QFL) classification continuation, with all other components frozen.

\subsection{Full Parameterizations}
\label{app:param}

\paragraph{Base transformation \(\mathcal B\).}
The Base token is a learnable transformation of the mean \(\mu_c\) computed with a learnable query \(q_B\) and feed-forward refinement; structurally, it is a single-seed PMA with uniform attention weights~\citep{lee2019settransformer}, i.e., a Deep-Sets-style encoding of the mean~\citep{zaheer2017deepsets}:
\begin{equation}
\tilde b_c = \LN(\mu_c+q_B),\qquad
B_c = \LN\!\left(\tilde b_c+\FFN_B(\tilde b_c)\right).
\label{eq:basefull}
\end{equation}

\paragraph{Residual slot update \(\mathcal R_r\).}
Each Residual slot \(r\) starts from a learnable query \(a_r\) and pools evidence deviations \(\delta_{cn}=e_{cn}-\sg(B_c)\). Attention logits are first normalized across slots so the slots compete for each evidence vector, and are then renormalized over valid evidence vectors, following the normalization pattern of Slot Attention~\citep{locatello2020slot}. Unlike Slot Attention, the slots pool deviations from a support-dependent anchor, use deterministic slot-specific queries rather than sampled initial states, and are updated by residual feed-forward steps rather than a GRU. After two iterations,
\begin{equation}
\alpha_{rn} =
\operatorname{softmax}_{r}\!\left(
\frac{\LN(a_r)^\top \LN(\delta_{cn})}{\sqrt d}
\right),\qquad
w_{rn}=\frac{\alpha_{rn}}{\sum_{n'}\alpha_{rn'}},
\label{eq:slotweightfull}
\end{equation}
\begin{equation}
a_r \leftarrow
\LN\!\left(a_r+W_u\sum_n w_{rn}\delta_{cn}\right),\qquad
a_r \leftarrow \LN\!\left(a_r+\FFN_s(a_r)\right),\qquad
R_{cr}=\LN(W_o a_r).
\label{eq:slotupdatefull}
\end{equation}
With a single Residual slot, the slot-wise softmax equals one and the weights are uniform, so \(R_{c1}\) is a learnable function of the offset \(\mu_c-B_c\) (Appendix~\ref{app:proofs}).

\paragraph{Set-reasoning block.}
\(\operatorname{SetBlock}\) applies one eight-head self-attention layer without index-dependent positional encoding, followed by a per-element \(256\!\rightarrow\!512\!\rightarrow\!256\) FFN, both with residual connections and LayerNorm:
\begin{equation}
\widetilde H = \LN\!\left(H^P+\MHA(H^P,H^P,H^P)\right),\qquad
H^F = \LN\!\left(\widetilde H+\FFN(\widetilde H)\right).
\label{eq:setblockfull}
\end{equation}

\section{Structural Properties and Proofs}
\label{app:proofs}

This section collects the formal statements of the structural properties discussed in Secs.~3.2 and~3.3. The properties themselves are elementary; their role is to make the design constraints explicit.

\paragraph{Proposition 1 (support-set encoding).}
For any \(K\ge 1\), \(\Phi_{\mathrm{BR}}(\mathcal S_c)\) contains \(R+1\) tokens, and for any permutation \(\pi\) of the \(K\) exemplars,
\[
\Phi_{\mathrm{BR}}(\pi\!\cdot\!\mathcal S_c)
=
\Phi_{\mathrm{BR}}(\mathcal S_c).
\]

\paragraph{Analysis: the default prompt is an anchor--correction factorization.}
Because attention logits are normalized across slots, the softmax over the single Residual slot of the default prompt (\(R=1\), T2) is identically one, so the Residual weights are uniform over valid evidence vectors and \(V_c=[B_c,R_{c1}]\) is a deterministic function of the mean \(\mu_c\), with \(R_{c1}\) a learnable correction computed from the offset \(\mu_c-B_c\). Although the default Base+R1 prompt therefore contains no information beyond the first-order pooled statistic \(\mu_c\), it is not equivalent to a raw mean prototype: it learns a nonlinear two-token re-parameterization \(F(\mu_c)=[B_c,R_{c1}]\) that maps the pooled visual evidence into a detector-compatible prompt space. What it cannot represent is higher-order structure of the support set---two support sets with identical means necessarily produce identical Base+R1 prompts; such distribution-sensitive routing becomes possible only when multiple Residual slots compete for individual evidence vectors (\(R\ge2\)). Section~4.3 shows that this extra capacity is unnecessary in the few-shot regime: with at most ten exemplars, Base+2R and Base+3R score no higher than Base+R1. The tested improvement over the mean-prototype controls therefore does not require instance-level multi-slot routing; in this few-shot setting, the simpler mean-dependent anchor--correction factorization is sufficient.

\paragraph{Proposition 2 (paired candidate sets).}
Assume both readers are permutation equivariant with respect to their queries, as in DETR-style decoders. If the initial queries are reordered by a permutation \(\pi\), then (i) the multiset of initialization pairs
\(\{(s_i^T,s_i^V)\}\) is unchanged, and (ii) \(\Psi_{\mathrm{PQA}}\) is permutation equivariant:
\[
\Psi_{\mathrm{PQA}}(\pi\!\cdot\!P)
=
\pi\!\cdot\!\Psi_{\mathrm{PQA}}(P),
\]
where \(P=[(s_i^T,s_i^V)]_{i=1}^{N}\).
Hence the final detection set is independent of query order except for score ties, and the Hungarian loss is also query-order invariant.

Pairing by initialization costs \(O(N)\). Aligning two independently decoded candidate sets would instead require choosing a cross-modal matching cost and solving an assignment problem, with \(O(N^3)\) complexity for the Hungarian algorithm. Proposition~2 also clarifies the role of the set module: pairwise arbitration is an elementwise equivariant map but cannot make one candidate depend on another, whereas the attention layer introduces exactly this interaction while preserving equivariance.

We use the notation from Section~3. A permutation \(\pi\) acts on a support set by reordering exemplars and on a list with \(N\) elements by reordering entries; \(\Pi\) denotes the corresponding permutation matrix.

\paragraph{Proof of Proposition 1.}
The instance encoder and the projection that produces evidence vectors use shared parameters and act independently on each exemplar. Reordering the exemplars therefore reorders the evidence vectors in groups of \(L\!\cdot\!G\). The mean \(\mu_c\) is invariant to any permutation of the evidence vectors, so \(B_c\) is also invariant. The Residual logits
\[
\frac{\LN(a_r)^\top\LN(\delta_{cn})}{\sqrt d}
\]
depend on \(n\) only through \(\delta_{cn}\). Reordering evidence vectors therefore permutes the columns of the slot-by-evidence logit matrix and permutes the padding mask in the same way. Normalization across slots is performed independently within each column, while the subsequent normalization over evidence vectors depends on the columns only through a sum over \(n\). Hence the pooled vector
\(\sum_n w_{rn}\delta_{cn}\) is invariant. Every slot update depends on the evidence only through these pooled vectors, so every iteration and every final \(R_{cr}\) is unchanged. By construction,
\[
V_c=[B_c,R_{c1},\ldots,R_{cR}]
\]
contains exactly \(R+1\) entries for any \(K\). \(\square\)

\paragraph{Proof of Proposition 2.}
(i) Reordering the initial queries by \(\pi\) and applying permutation-equivariant readers places \(s_{\pi(i)}^T\) and \(s_{\pi(i)}^V\) at position \(i\). The pair at position \(i\) is therefore
\((s_{\pi(i)}^T,s_{\pi(i)}^V)\), so the multiset of pairs is unchanged.

(ii) The arbitrator uses shared parameters and computes \(x_i\), \(g_i\), \(h_i^P\), and \(r_i^P\) only from pair \(i\), so it commutes with \(\pi\). Self-attention without index-dependent positional encoding satisfies
\[
\MHA(\Pi H,\Pi H,\Pi H)
=
\Pi\,\MHA(H,H,H),
\]
because the attention-logit matrix transforms as \(\Pi M\Pi^\top\), row-wise softmax commutes with this simultaneous row/column permutation, and \(\Pi^\top\Pi=I\). LayerNorm, FFN, and the prediction head act independently on each element. The composition of equivariant maps remains equivariant, so the prediction head outputs the same detection multiset regardless of query order. The Hungarian loss, which minimizes over all assignments, also retains the same value. \(\square\)

\end{document}